\documentclass[10pt, a4paper]{article}
\usepackage{lrec}
\usepackage{multibib}
\usepackage{graphicx}
\usepackage{tabularx}
\usepackage{soul}
\usepackage{pgfplots}
\usepackage{xcolor}
\usepackage{epstopdf}
\usepackage[latin1]{inputenc}
\usepackage{hyperref}
\usepackage{xstring}
\usepackage{relsize}
\usepackage{amssymb}
\usepackage{mdframed}
\usepackage{amsmath}
\pgfplotsset{compat=1.14}

\usetikzlibrary{patterns}

\title{Automated Evaluation of Out-of-Context Errors}

\name{Patrick Huber, Jan Niehues, Alex Waibel}

\address{Institute for Anthropomatics and Robotics, Karlsruhe Institute of Technology (KIT)\\
uhejt{@}student.kit.edu, jan.niehues{@}kit.edu, alex.waibel{@}kit.edu\\
}

\abstract{
\begin{center}
We present a new approach to evaluate computational models for the task of text understanding by the means of out-of-context error detection. Through the novel design of our automated modification process, existing large-scale data sources can be adopted for a vast number of text understanding tasks. The data is thereby altered on a semantic level, allowing models to be tested against a challenging set of modified text passages that require to comprise a broader narrative discourse.\\
Our newly introduced task targets actual real-world problems of transcription and translation systems by inserting authentic out-of-context errors. The automated modification process is applied to the 2016 TEDTalk corpus. Entirely automating the process allows the adoption of complete datasets at low cost, facilitating supervised learning procedures and deeper networks to be trained and tested. To evaluate the quality of the modification algorithm a language model and a supervised binary classification model are trained and tested on the altered dataset. A human baseline evaluation is examined to compare the results with human performance. The outcome of the evaluation task indicates the difficulty to detect semantic errors for machine-learning algorithms and humans, showing that the errors cannot be identified when limited to a single sentence.
\end{center}
}
\begin{document}
\maketitleabstract
\section{Introduction}
Machine learning strives to achieve universal application in solving arbitrary tasks rather than special functionalities and well-defined functions. This trend is reflected by the increasing importance of text understanding within the field of natural language processing (NLP).\\
Extracting the context of a narrative passage can thereby benefit a wide variety of applications, such as automatic speech recognition (ASR) and neural machine translation (NMT) tasks. Most state-of-the-art ASR and NMT systems are currently only processing one sentence at a time, solely basing the system's decisions on the local context \cite{Bahdanau-Cho-Bengio-14,Cho-et-al-14}. This approach is reasonable, as it addresses the most general case. Nevertheless, considering a larger context can highly benefit the overall system's performance for text passages with contextual features. \\
Imagine an ASR system that is supposed to transcribe the sentence \textit{There are great opportunities in the far-east} within a presentation on the global economy. Through an inarticulate pronunciation and a noisy signal at the conference, the ASR system transcribes the sentence as \textit{There are great opportunities in the forest}, as the sentences are phonetically very similar. In this example, the language model of the ASR system will not be able to identify the out-of-context error by only considering the current sentence.\\
Therefore, developing a context component to enhance the text understanding is an important task to improve state-of-the-art systems.\\
While the number of published datasets regarding question-answering (QA) tasks steadily increases, for example based on children books \cite{Hill-Bordes-Chopra-Weston-15} or simple reasoning \cite{Weston-Bordes-Chopra-Mikolov-15}, there are very limited resources covering the area of context-aware error detection based on a narrative.\\
For this reason, we designed a novel text understanding task, which is difficult to solve by only relying on the local context of a sentence, but becomes feasible when taking a broader discourse into account. We introduce a fully automated dataset extension procedure, designed to assess the performance of computational models to identify out-of-context errors. Through the fully automated modification of the data, the out-of-context errors vary within a complexity range. The diversity of the modification severity generates a broad scale to compare computational approaches.\\
To evaluate the performance of the modification procedure, we apply the process on the 2016 TEDTalk corpus \cite{Cettolo-Girardi-Federico-12} and test multiple baseline systems against the dataset.\\
Therefore, a standard sentence-based language model (LM) is evaluated on the corpus, reaching a F-score of 6.51\%. To show the performance of supervised models, a binary classification network is set up and trained on the modified corpus achieving a F-score of 10.16\% on the test set. To gain additional insight on the relative performance of the computational baseline models, we conduced a human baseline survey, showing the difficulty of the task for human subjects (see section \ref{Evaluation} for more details).
\section{Related Work}
The LAMBADA dataset introduced by \newcite{Paperno-et-al-16} is one of the first and most comprehensive data sources addressing the task of text understanding by the means of word prediction. Through a multi-staged filtering approach and extensive evaluation of each text passage, the corpus represents a high quality data source for context-aware word prediction tasks.

Nevertheless, generating these handcrafted data sources is very expensive and thus limited. For example, to be able to train supervised models, not only does the relatively small development and test sets need to be processed, but also the large set of training data.

Our newly developed alteration approach addresses this downside through the application of a fully automated modification procedure.

Even though the LAMBADA dataset and our newly introduced modification process are targeting the same overall task, the performed inference on the data varies. The LAMBADA inference task targets the continuation of the current text passage at a known position in the text by predicting the last word within the paragraph. In contrast to that, our newly introduced substitution process replaces context-relevant words within the text passage at arbitrary positions, increasing the complexity of the task.

\newcite{Sennrich-16} introduces a dataset with automatically inserted errors focusing on advanced computational models for NMT tasks.

The paper by \newcite{Burlot-Yvon-17} proposes an evaluation process for NMT models, assessing the morphological properties of a system. The process substitutes nouns, as well as other part-of-speech tokens with filtered and randomly chosen replacement words.
\section{Task}
\label{task}
The task introduced in this paper is designed to evaluate the performance of computational models for out-of-context error detections.\\
The fully automated modification process described in section \ref{Data_modification} provides the ground truth for the task. 
The artificially inserted out-of-context tokens are uniformly distributed over the dataset, elevating the complexity of the task over fixed-position approaches. 
Compared to approaches with well-known target positions of substituted words $w\textsubscript{n}$ (e.g. at the end of a paragraph), the out-of-context word replacements within this task are at random positions $w\textsubscript{p}$.
The series of word-tokens is thereby described by the ordered sequence of words $W = (w\textsubscript{1}, w\textsubscript{2}, ..., w\textsubscript{n})$ with $w\textsubscript{p} \in W$.\\
The task to find the correct word $w\textsubscript{n}$ at a known position (in this case at the end of the paragraph of length $N$) can be described as a classification task with one class per word in the vocabulary $m\textsubscript{vocab}$.
The assumption to know the position of the replaced out-of-context word $w\textsubscript{p}$ does not hold, as the substitutions are randomly distributed. Instead, every word $w\textsubscript{p} \in W$ needs to be assessed against every other word $w\textsubscript{q} \in W$ in the sequence with $p \neq q$.\\
The presented task can therefore be interpreted as a binary sequence labeling problem defined by the input sequence $W = (w\textsubscript{1}, w\textsubscript{2}, ..., w\textsubscript{n})$ of length $N$ representing the text passage and the output sequence $L\textsubscript{OOC} = (l\textsubscript{1}, l\textsubscript{2}, ..., l\textsubscript{n})$, also of length $N$, with $l\textsubscript{i} \in L\textsubscript{OOC}$ representing the label of the input element $w\textsubscript{i} \in W$. The labels $L\textsubscript{OOC}$ thereby separate the two classes $\{0, 1\}$, representing valid-context tokens (0) and out-of-context tokens (1).\\
The labeled text passage:\\

$W \cup L\textsubscript{OOC}$ = (\textit{(We, 0) (have, 0), (in, 0) (higher, 0) (education, 0), (a, 0) (trillion, 0) (dollars, 0) (of, 0) (student, 0) (debt, 0) [...]. (We, 0) (have, 0) (a, 0) (lot, 0) (of, 0) \textbf{(shopping, 1)}. (Our, 0) (economy, 0) (grew, 0) [...] (on, 0) (the, 0) (back, 0) (of, 0) (consumers, 0) (massively, 0) (over, 0) (borrowing, 0).})

from the presentation \textit{The death of innovation, the end of growth} by Robert Gordon, held in February 2013, gives an example on how the binary sequence labeling $L\textsubscript{OOC}$ can be applied.\\
Our task definition does not provide any information about the position of the modifications, nor give any insight about the total number of out-of-context substitutions on the data. \\
To identify the dataset replacements with a computational model, the narrative of the text passage needs to be employed. A context $C$ is thereby defined as a coherent text passage of up to ten sentences containing multiple appearances of the same reoccurring noun $n$.
\section{Data Modification}
\label{Data_modification}
The goal of the dataset manipulation is to modify an existing database by artificially inserting out-of-context errors in text passages with especially strong contextual features. The substitution procedure is fully automated to enable the modification of entire large-scale datasets without the costly validation of the data by human subjects. A crucial task emerging through the automated processing is the reasonable replacement of context related words. The substitutions fulfill the following two requirements: \\
First of all, the modified dataset serves as the ground truth to test the train context-aware computational models against.\\
Secondly, through the out-of-context modifications on the training set, supervised models can be trained on the data.\\
This is achieved through a four-stage computational procedure:\\

(1) Dataset Filtering\\
To enhance the dataset quality, non-contextual parts of the data are removed, excluding self-contained short text passages with less than 200 words. \\

(2) Part-of-Speech Tagging \\
For the semantic out-of-context substitutions, only certain part-of-speech (POS) classes are taken into account. Thus, every token in the dataset needs to be assigned a POS category. As described by \newcite{Paperno-et-al-16}, context is especially critical to nouns, whereas other POS classes can often be inferred directly out of the local context of a sentence. Our substitution process therefore focuses on the replacement of nouns.\\

(3) Candidate Selection\\
To ensure that a text passage contains a sufficient context, the nouns determined by the POS-tagger are filtered for contextual coherence. A context is thereby assumed if the same noun appears multiple times within the same text passage. The last appearance of the noun in a context of ten consecutive nouns qualifies as a suitable replacement and is saved as a potential out-of-context substitution candidate.\\
Out of the list of potential replacement candidates, a predetermined number of tokens is randomly selected according to a uniform distribution.\\
\\

(4) Appearance Window\\
For every selected token in the original dataset, a syntactically suited replacement token is determined. The tokens are thereby replaced by words within a defined appearance window. The appearance window approach is based on the assumption, that words with a similar word count on the original dataset are suitable replacements. The approach has multiple characteristics that enhance the quality of substitutions: (a) A similar word count of replacements avoids common words being substituted by rare words. (b) The replaced words are typically not related to the original word, as the overall word count generally not infers semantic affiliation.\\

Within the appearance window further filtering regarding the tense and grammatical number are executed to determine the most suitable substitution.\\
This process can be applied to arbitrary data sources with contextual features to train and test context-aware models on. To show the results of the modification process and evaluate the quality of the semantic replacement tokens, we apply the modification procedure on the 2016 TEDTalk corpus\footnote{https://github.com/isl-mt/SemanticWordReplacement-LREC2018} and assess the semantic out-of-context dataset substitutions on supervised as well as unsupervised baseline models.\\
\section{Baselines}
\label{Baseline_Model}
To evaluate the performance of the modification process (section \ref{Data_modification}) on the newly introduced task (section \ref{task}), three baselines are designed. The baseline neural LM evaluates the quality of the modification pipeline for unsupervised models. A standard binary classification neural network assesses the performance of supervised models. A human baseline is additionally acquired to assess the human performance on the task.\\

(1) Language Model\\
The architecture of the baseline neural LM is derived from the original model proposed by \newcite{Bengio-et-al-03}, which employs a fundamental architecture for the neural network. The basic recurrent neural network (RNN) architecture is extended by state-of-the-art LSTM units to enhance the model's performance on the task, as proven more effective by \newcite{Xie-Rastogi-17} and \newcite{Sundermeyer-et-al-12}. The shallow design of the standard baseline LM to evaluate the quality of the modification process contains three computational layers.\\
The embedding layer is trained on the complete input of the neural network and encodes the sparse word representations into 256-dimensional real-valued word embedding vectors.\\
The 256-dimensional word embeddings are subsequently fed into the 512 LSTM units within the recurrent network layer.\\
To expose the outputs of the LSTM units, the output layer comprises of one computational unit per class. In order to normalize the output of the network, a softmax activation function is chosen as the computation of the output layer.\\
The model is trained with the Adam optimizer, a learning rate of 1e-3 and the cross entropy as the selected cost function. The size of the model is limited to a vocabulary size of 30,000 words and a maximum sequence length of 50 words.\\

(2) Binary Classification Model\\
The binary classification model is the supervised baseline model to evaluate the modification process against the defined task. In order to keep the approaches comparable, the architecture and hyper-parameters of the supervised baseline network are selected to be similar to the design of the unsupervised neural LM.\\
The main difference between the models is the output layer, as the binary classification model only distinguishes two classes $\{0, 1\}$. All other properties of the systems are identical.\\

(3) Human Baseline\\
To assess the human baseline performance on the newly introduced task, ten random sentences with one or more replacement tokens are selected from the modified dataset and presented to seven human participants in random order. The task description for the human baseline was to find the word(s) in the sentences that do not fit the context. With the inserted replacements being intentionally designed to be semantically rather that syntactical, the replacements are deliberately difficult to identify with only one sentence.\\
To be able to directly compare the human results to the two baseline neural networks, the computational models are also exclusively tested on the ten sentences (232 words) utilized for the human baseline evaluation.

\section{Evaluation}
\label{Evaluation}
The modification process (section \ref{Data_modification}) is applied and evaluated on the English TEDTalk corpus released for the IWSLT 2016 \cite{Cettolo-Girardi-Federico-12}. The original TEDTalk database contains transcripts of over 2,600 TEDTalks presented between 2007 and 2016. With a well-defined topic per TEDTalk and an average duration of 30 to 45 minutes per presentation, the TEDtalk corpus contains strong contextual features within long and coherent contexts. Furthermore, the TEDtalk dataset covers a wide variety of topics keeping the data mostly unbiased.\\
After applying the replacement procedure on the original TEDTalk dataset, over 25,000 contextual words are replaced by out-of-context tokens. The following example is taken from the modified TEDTalk corpus to illustrate the result of the substitution process.\\

Local context: \textit{And that's the world we're coming into, in which we will increasingly see that our \textbf{marketers} are not fixed.}\\

Extended context: \textit{Now notice, in a period which is dominated by a mono-polar world, you have fixed alliances -- NATO, the Warsaw Pact. A fixed polarity of power means fixed alliances. But a multiple polarity of power means shifting and changing alliances. And that's the world we're coming into, in which we will increasingly see that our \textbf{marketers} are not fixed.}\\

Ground truth: \textit{marketers $\rightarrow$ alliances}\\

(Paddy Ashdown, December 2011 at TEDxBrussels)\\

\begin{figure}[!h]
\begin{center}
\begin{tikzpicture}
\begin{axis}[ybar,enlargelimits=0.15,legend pos=north west,ylabel={\% in Quarters},symbolic x coords={0-25\%,25-50\%,50-75\%,75-100\%},xtick=data]
\addplot[fill=cyan, postaction={pattern=dots}] coordinates {(0-25\%,1.95) (25-50\%,1.25) (50-75\%,2.67) (75-100\%,93.35)};
\addplot[fill=purple, postaction={pattern=horizontal lines}] coordinates {(0-25\%,2.2) (25-50\%,2.45) (50-75\%,9.95) (75-100\%,84.6)};
\legend{Original TEDTalk Tokens, Replaced TEDTalk Tokens}
\end{axis}
\end{tikzpicture}
\caption{Substitution Quality of Modified Tokens}
\label{figure_on_spot}
\end{center}
\end{figure}
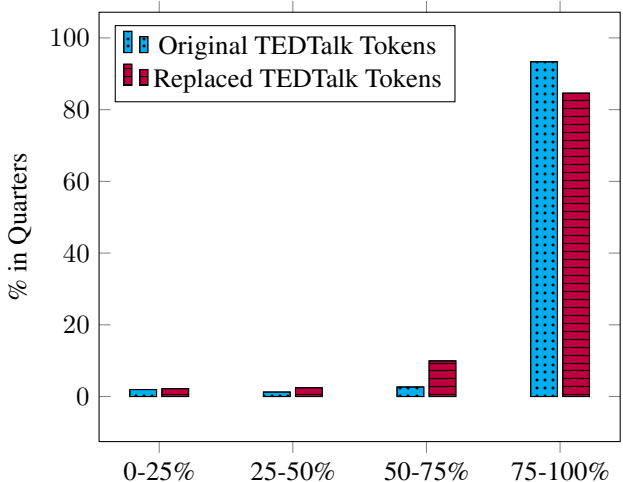

The substituted words in the text passages shown above are difficult to identify by only relying on the local context of the sentence. However, taking the broader discourse of the extended context into account discloses which words are out of context. \\
As the modifications on the TEDTalk dataset are fully automated, not all replacements are equally challenging. Within the complexity range of modifications, the replacements vary from easy substitutions, which can eventually be identified by only assessing the local context, to difficult substitution, which are hard to infer for human subjects. Through the design of the processing pipeline, the number of these outliers is minimized.\\
Figure \ref{figure_on_spot} supports this assertion by comparing the distribution of the original TEDTalk tokens with the replaced words using the computational baseline model described in section \ref{Baseline_Model}. Thereby, every word in the vocabulary m\textsubscript{vocab} is assessed at the replaced positions. The probabilities of the words are sorted and divided into four equal quarters representing the four classes on the x-axis. The distribution of the original TEDTalk tokens with 93.35\% of the words within the top 25\% of the vocabulary shows the effectiveness of the baseline LM on the original data. The distribution of the modified words, with 84.6\% of the artificially replaced words within the highest quarter (75\%-100\%), shows that the replacements also fit well in the local context.\\

To further assess the quality of the replacements, we evaluate the results of the modification process by training and testing the baseline models described in section \ref{Baseline_Model}.\\
As the baseline LM is unsupervised, the unmodified TEDTalk dataset is used to train the computational baseline model. The performance is tested against the modified test set.\\
To compare the performance of the computational models, the F-score is assessed. As described in section \ref{task}, the absolute amount of replacements, as well as the positions of the out-of-context error tokens is not provided. Therefore, to assess the quality of the model, the probability of every word within the dataset $D$ is calculated. The generated set of probabilities is subsequently sorted to classify the tokens with the lowest probability $N\textsubscript{ooc}$ as out-of-context tokens.
With the dynamic threshold $t\textsubscript{N\textsubscript{ooc}} = \{1 .. N\}$ with $ 1 \le N \le |D|$ to separate
the binary classes, the best F-score is retrieved.\\
For the unsupervised model, the perplexity measure is additionally
applied to show the systems' performance on the general word prediction task.\\
The results are shown in table \ref{tab:table_results}. The reported performance for the unsupervised neural LM on the test set is a perplexity value of 115 and a F-score of 6.51\%.\\

\begin{table}[!h]
\begin{center}
\begin{tabular}{ l | l | l | l}
Model & Perplex & F-Score & F-score*\\
\hline
Lang. Model & 115 & 6.51\% & 8.65\%\\
Bin. Class. Model & - & 10.16\% & 13.43\%\\
Human Survey & - & - & 28.53\%\\
\end{tabular}
\caption{Out-of-Context Detection Rates on the Test-Set (F-score* refers to a randomly chosen subset of the test set to compare the computational and human baselines)}
\label{tab:table_results}
\end{center}
\end{table}

The second baseline model is the supervised binary classification model. This model particular benefits from the automated manipulation process, as the training and the testing require tagged data. In order to separate the two classes $\{0, 1\}$, the model learns from the labeled ground-truth data. As the positions of the replaced out-of-context tokens are not known, the separation between the classes is solely based on the score of the tokens. To divide the model's output into the two classes, it is sorted by the probability of the words and separated at the best threshold. The best result achieved by the model is a F-score of 10.16\%.\\

As the third score, a human evaluation has been implemented\footnote{https://github.com/isl-mt/SemanticWordReplacement-LREC2018}. This way, we can compare the computational results with the human performance.\\
The displayed F-score* in Figure \ref{tab:table_results} shows the direct comparison between the baseline LM, the baseline binary classification model and the results of the human survey on a randomly selected subset of the test data. 

Both baseline models thereby achieve comparable results on the randomly chosen subset as to the score on the complete test set (8.65\% for the LM and 13.43\% for the binary classification model), indicating that the subset is representative. The human baseline evaluation shows a better result, reaching an F-score of 28.53\%.\\
This outcome indicates that the presented task is not only difficult to solve for computational models, but also for humans, when the context of a sentence is not available. If the user is able to see an extended context on the other hand, it is nearly always possible to find the error, as shown in some preliminary analysis.\\
With this result, we additionally show the integrity of the baseline models, performing on a reasonable level compared to the human performance. 

\section{Conclusion}
This work presents a novel task to evaluate context-aware computational models. The task reproduces common errors of state-of-the-art transcription and translation systems 
that can be prevented by taking a broader discourse into account. The arbitrary positions of the inserted errors classify the task within the area of sequence labeling. \\
Our newly introduced substitution approach to modify existing datasets with out-of-context tokens represents the ground-truth for the assessment of the task. We show that the substituted tokens are difficult to infer by computational models that solely employ the local context. Thus, a context-aware approach needs to be developed to enhance the performance on the task.\\
The newly defined task is a comprehensive benchmark for context-aware models to train and test the ability of the system to abstract from the syntactical noise of the text passage and learn to focus on the semantic representation of the narrative. Through our automated modification pipeline, the necessary ground-truth generation is accessible and fast.\\
Therefore, the introduced combination of a real-world task definition and an automated processing pipeline represents a complete framework to test future models on the task of text understanding by the means of out-of-context error detection.\\

\section{Bibliographical References}
\label{main:ref}

\bibliographystyle{lrec}
\bibliography{references}

\end{document}